\documentclass[letterpaper]{article} % DO NOT CHANGE THIS
\usepackage{aaai24}  % DO NOT CHANGE THIS
\usepackage{times}  % DO NOT CHANGE THIS
\usepackage{helvet}  % DO NOT CHANGE THIS
\usepackage{courier}  % DO NOT CHANGE THIS
\usepackage[hyphens]{url}  % DO NOT CHANGE THIS
\usepackage{graphicx} % DO NOT CHANGE THIS
\usepackage{natbib}  % DO NOT CHANGE THIS AND DO NOT ADD ANY OPTIONS TO IT
\usepackage{caption} % DO NOT CHANGE THIS AND DO NOT ADD ANY OPTIONS TO IT
\usepackage{algorithm}
\usepackage{algorithmic}

\newtheorem{theorem}{Theorem}
\newtheorem{lemma}{Lemma}

\usepackage{subfigure}
\usepackage{amsmath}
\usepackage{amsfonts}
\usepackage{booktabs}

\usepackage{newfloat}
\usepackage{listings}
\DeclareCaptionStyle{ruled}{labelfont=normalfont,labelsep=colon,strut=off} % DO NOT CHANGE THIS
\floatstyle{ruled}
\newfloat{listing}{tb}{lst}{}
\floatname{listing}{Listing}
\title{Improving Expressive Power of Spectral Graph Neural Networks\\ 
with Eigenvalue Correction}

\author{
    Kangkang Lu\textsuperscript{\rm 1}, Yanhua Yu\textsuperscript{\rm 1,}\thanks{Corresponding author.}, Hao Fei\textsuperscript{\rm 2}, Xuan Li\textsuperscript{\rm 1}, Zixuan Yang\textsuperscript{\rm 1}, Zirui Guo\textsuperscript{\rm 1}, Meiyu Liang\textsuperscript{\rm 1}, Mengran Yin\textsuperscript{\rm 1}, Tat-Seng Chua\textsuperscript{\rm 2}
}

\affiliations{
    \textsuperscript{\rm 1}Beijing University of Posts and Telecommunications \\ \textsuperscript{\rm 2}National University of Singapore\\
    
    \{lukangkang, yuyanhua\}@bupt.edu.cn, haofei37@nus.edu.sg, lixuan20000530@bupt.edu.cn,\{zxyang.bupt, zrguo.bupt\}@qq.com, \{meiyu1210, 2021110996\}@bupt.edu.cn, dcscts@nus.edu.sg
}

\usepackage{bibentry}
\begin{document}

\maketitle

\begin{abstract}
In recent years, spectral graph neural networks, characterized by polynomial filters, have garnered increasing attention and have achieved remarkable performance in tasks such as node classification. These models typically assume that eigenvalues for the normalized Laplacian matrix are distinct from each other, thus expecting a polynomial filter to have a high fitting ability. However, this paper empirically observes that normalized Laplacian matrices frequently possess repeated eigenvalues. Moreover, we theoretically establish that the number of distinguishable eigenvalues plays a pivotal role in determining the expressive power of spectral graph neural networks. In light of this observation, we propose an eigenvalue correction strategy that can free polynomial filters from the constraints of repeated eigenvalue inputs. Concretely, the proposed eigenvalue correction strategy enhances the uniform distribution of eigenvalues, thus mitigating repeated eigenvalues, and improving the fitting capacity and expressive power of polynomial filters. Extensive experimental results on both synthetic and real-world datasets demonstrate the superiority of our method. The code is available at: https://github.com/Lukangkang123/EC-GNN
\end{abstract}

\section{Introduction}

\begin{figure}[t] 
	\centering
	\subfigure[Cora]{\includegraphics[width=0.48\linewidth]{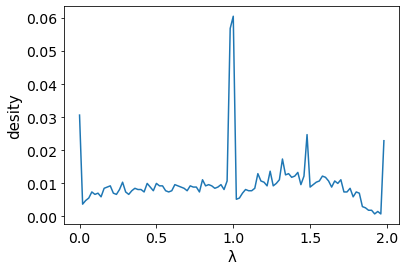}}
	\subfigure[Photo]{\includegraphics[width=0.48\linewidth]{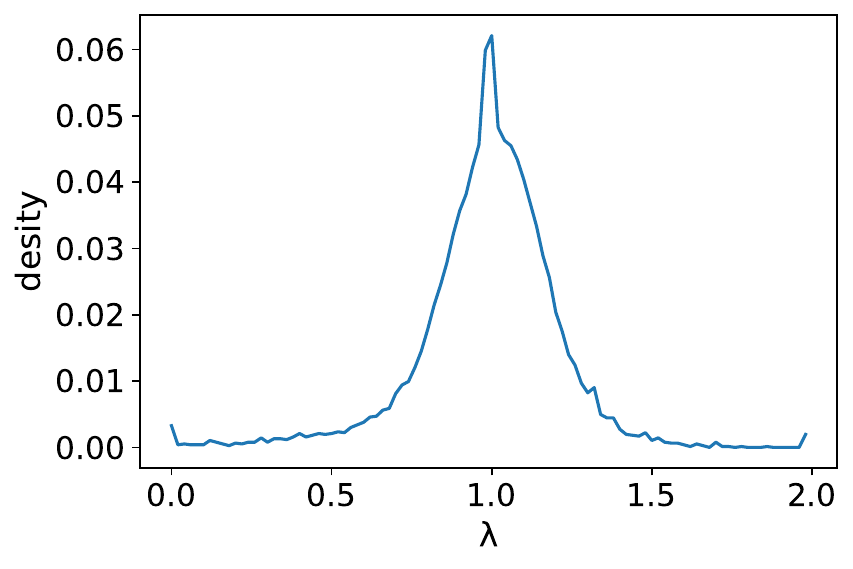}}
        \subfigure[Chameleon]{\includegraphics[width=0.48\linewidth]{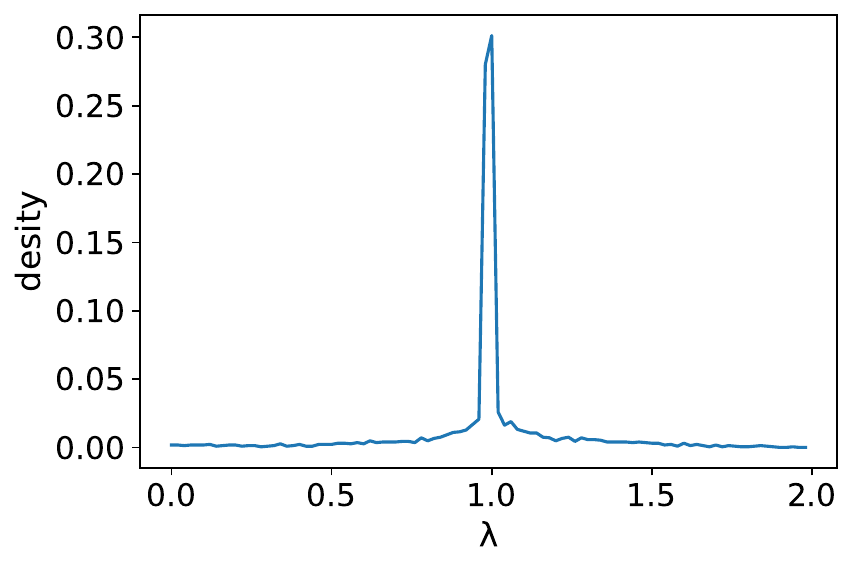}}
	\subfigure[Squirrel]{\includegraphics[width=0.48\linewidth]{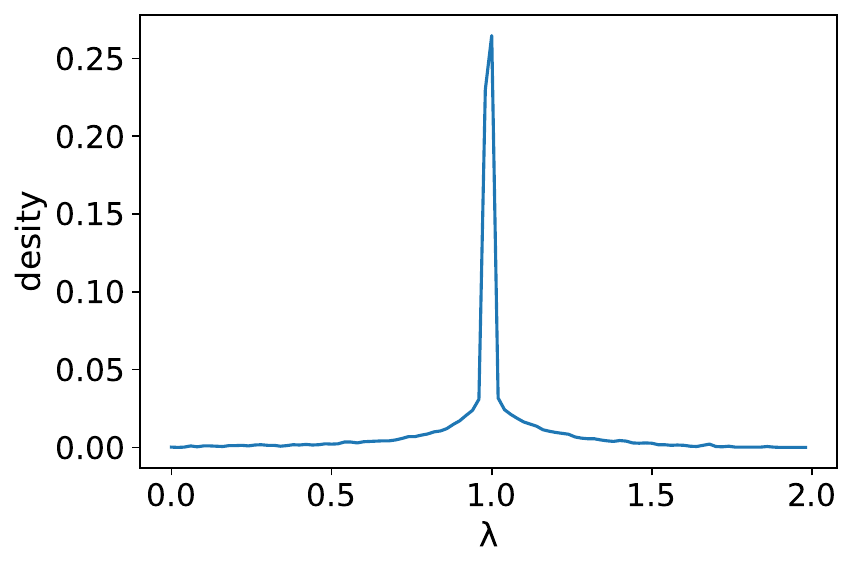}}
	\caption{The distribution of eigenvalues of the normalized Laplacian matrix on different datasets. The abscissa represents eigenvalues, and the ordinate represents the probability density.}
	\label{fig:histogram}
 % \vspace{-4mm}
\end{figure}

Graph neural networks (GNNs) have recently achieved excellent performance in graph learning tasks such as abnormal detection \cite{gaoaddressing,gaoalleviating}, molecular prediction \cite{fangjf2moltc,OAR,CPUTP}, and recommendation systems \cite{GNN_recommendation1,MENSA}. 
Existing GNNs can be divided into two categories: \emph{spatial GNNs} and \emph{spectral GNNs}. 
Spatial GNNs often adopt a message-passing mechanism to learn node representations by aggregating neighbor node features. Spectral GNNs map node features to a new desired space by selectively attenuating or amplifying the Fourier coefficients induced by the normalized Laplacian matrix of the graph. This study primarily centers on the realm of spectral GNNs.

\begin{figure}[t] 
	\centering
	\subfigure[seven eigenvalues]{\includegraphics[width=0.49\linewidth]{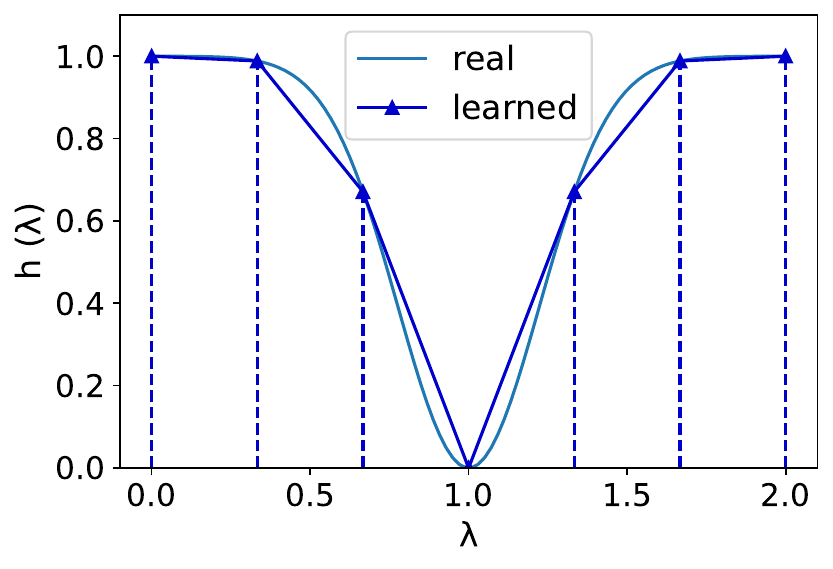}}
	\subfigure[eleven eigenvalues]{\includegraphics[width=0.49\linewidth]{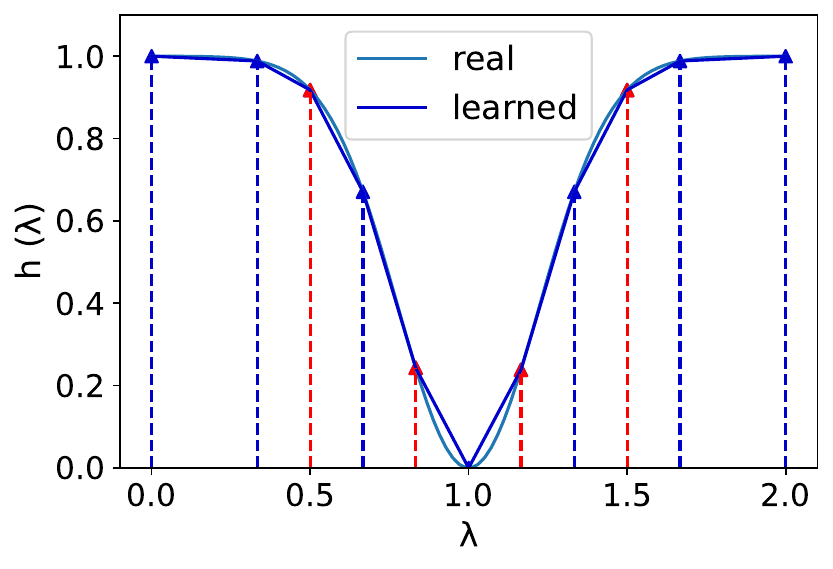}}
      % \subfigure[cora\_e\_new\_e]{\includegraphics[width=0.6\linewidth]{images/cora_e_new_e.png}}
	\caption{Illustration of two band-rejection filters with different numbers of eigenvalues. The red dotted line in (b) represents more eigenvalues than (a).}
	\label{fig:band_rejection}
 % \vspace{-4mm}
\end{figure}

Existing spectral filters are commonly approximated using fixed-order (or truncated) polynomials. Specifically, these polynomial filters take the normalized Laplacian eigenvalues as input and map them to varying scalar values via a polynomial function. However, our empirical findings unveil that the eigenvalues of normalized Laplacian matrix in real-world graphs tend to exhibit high multiplicity. This phenomenon is visually demonstrated in Figure 1\footnote{Additional distribution histograms for more datasets are provided in Appendix B.}, where a substantial number of eigenvalues cluster around the value 1. Consequently, when two frequency components share the same eigenvalue $\lambda$, they will be scaled by the same amount $h(\lambda)$\cite{JacobiConv}. As a result, the coefficients of these frequency components will retain the same proportion in predictions as in the input. And an abundance of repeated eigenvalues inevitably curbs the expressive capacity of polynomial filters. To illustrate, Figure \ref{fig:band_rejection} highlights how a filter possessing seven distinct eigenvalues exhibits weaker fitting capabilities than the one endowed with eleven distinct eigenvalues.

Based on the insights gained thus far, it's apparent that the eigenvalue multiplicity of the normalized Laplacian matrix significantly impedes the polynomial filter's capacity to effectively fit. 
More severely, the challenge cannot be readily addressed by simply elevating the polynomial order, because even exceedingly high-order polynomials could fail to differentiate outputs for identical eigenvalues. 
To tackle this, we propose an \textbf{eigenvalue correction strategy} to generate more distinct eigenvalues, which reduces eigenvalue multiplicity and in return enhances the fitting prowess of polynomial filters across intricate and diverse scenarios.
In essence, our strategy combines the original eigenvalues and the eigenvalues sampled at equal intervals from [0, 2].
This amalgamation ensures a more uniform eigenvalue distribution while still retaining the frequency information encapsulated within the eigenvalues. 
Furthermore, we manage to enhance training efficiency without excessively elongating precomputation time, which allows us to efficiently amplify the order of polynomial spectral GNNs, contributing to a refined and potent modeling capability.

To summarize, our contributions in this work are highlighted as follows:

\begin{itemize}
\item We empirically discover that the normalized Laplacian matrix of a majority of real-world graphs exhibits elevated eigenvalue multiplicity. 
Furthermore, we provide theoretical evidence that the distinctiveness of eigenvalues within the normalized Laplacian matrix critically shapes the efficacy of polynomial filters. 
Consequently, a surplus of repeated eigenvalues obstructs the potential expressiveness of these filters.

\item We introduce an innovative eigenvalue correction strategy capable of substantially diminishing eigenvalue multiplicity. 
By ingeniously amalgamating original eigenvalues with equidistantly sampled counterparts, our method can seamlessly integrate with existing polynomial filters of GNNs.

\item Extensive experiments on synthetic and real-world datasets unequivocally demonstrate the superiority of our proposed approach over state-of-the-art polynomial filters. 
This substantiates the remarkable enhancement in the fitting prowess and expressive capacity of polynomial filters, achieved through the novel eigenvalue correction strategy we present.

\end{itemize}

\section{Related Work}
\subsection{Spectral GNNs}
According to whether the filter can be learned, the spectral GNNs can be divided into \emph{predefined filters} \cite{GCN,PPNP,GNN_LF_HF} and \emph{learnable filters} \cite{Chebyshev,GPR-GNN,BernNet,JacobiConv}. 

In the category of predefined filters, GCN \cite{GCN} uses a simplified first-order Chebyshev polynomial, which is shown to be a low-pass filter. APPNP \cite{PPNP} utilizes Personalized Page Rank (PPR) to set the weight of the filter. GNN-LF/HF \cite{GNN_LF_HF} designs filter weights from the perspective of graph optimization functions, which can fit high-pass and low-pass filters.

Another category of spectral GNNs employs learnable filters. ChebNet \cite{Chebyshev} uses Chebyshev polynomials with learnable coefficients. GPR-GNN \cite{GPR-GNN} extends APPNP by directly parameterizing its weights and training them with gradient descent. ARMA \cite{ARMA} learns a reasonable filter through an autoregressive moving average filter family. BernNet \cite{BernNet} uses Bernstein polynomials to learn filters and forces all coefficients positive. JacobiConv \cite{JacobiConv} adopts an orthogonal and flexible Jacobi basis to accommodate a wide range of weight functions. FavardGNN \cite{FavardGNN} learns an optimal polynomial basis from the space of all possible orthonormal basis.

However, the above filters typically use repeated eigenvalues as input, which hinders their expressiveness. In contrast, we propose a novel eigenvalue correction strategy that can generate more distinct eigenvalues and thus improve the expressive power of polynomial filters.

\begin{table*}[t]
\begin{tabular}{ccccccccccc}
\hline
datasets   & Cora   & Citeseer & Pubmed & Computers & Photo & Texas  & Cornell & Squirrel & Chameleon & Actor   \\
\hline
$\|\mathcal{V}\|$   & 2708   & 3327  & 19717  & 13752  & 7650 & 183  & 183 & 5201  & 2277 & 7600   \\
$\|\mathcal{E}\|$    & 5278   & 4552     & 44324  & 245861  & 119081  & 279   & 277   & 198353   & 31371     & 26659  \\
$N_{\text{distinct}}$  & 2199   & 1886   & 7595   & 13344  & 7474   & 106    & 115    & 3272   & 1120   & 6419   \\
$P_{\text{distinct}}$& 81.2\% & 56.7\% & 38.5\% & 97.0\% & 97.7\% & 57.9\% & 62.8\% & 62.9\% & 49.2\% & 84.5\% \\
\hline
\end{tabular}
\caption{Dataset statistics. $\|\mathcal{V}\|$  is the number of nodes, $\|\mathcal{E}\|$ is the number of edges, $N_{\text{distinct}}$ is the number of distinct eigenvalues, and $P_{\text{distinct}}$ is the proportion of distinct eigenvalues to all eigenvalues.
}
\label{tab:datasets}
\end{table*}

\subsection{Expressive Power of Spectral GNNs}
The Weisfeiler-Lehman test \cite{WL_test} is considered as a classic algorithm for judging graph isomorphism. GIN \cite{GIN} shows that the 1-WL test limits the expressive power of GNNs to distinguish non-isomorphic graphs. \citet{gnn-spectral-expressive-power} theoretically demonstrates some equivalence of graph convolution procedures by bridging the gap between spectral and spatial designs of graph convolutions. JacobiConv \cite{JacobiConv} points out that the condition for spectral GNN to generate arbitrary one-dimensional predictions is that the normalized Laplacian matrix has no repeated eigenvalues and no missing frequency components of node features.

\subsection{Distribution of Eigenvalues of Normalized Laplacian Matrix}
For random graphs, \citet{spectrum_empirical_distributions} proves that the empirical distribution of the eigenvalues of the normalized Laplacian matrix of Erdős-Rényi random graphs converges to the semicircle law.  \citet{Spectra_random} also proves that the Laplacian spectrum of a random graph with expected degree obeys the semicircle law. Empirically, we observe that as the average degree of a random graph increases, its eigenvalue distribution becomes more concentrated around 1. Histograms of their distributions can be found in Appendix C.

For real-world graphs, we empirically find that almost all real-world graphs have many repeated eigenvalues, especially around the value 1, which is consistent with the findings of QPGCN \cite{QPGCN} and SignNet \cite{SignNet}. Furthermore,  some works \cite{motif_duplication1,motif_duplication2} attempt to explain the generation of the multiplicity of eigenvalue 1 by motif doubling and attachment. The number and proportion of distinct eigenvalues for real-world graphs are shown in Table \ref{tab:datasets}. It's evident that almost all datasets do not have enough distinct eigenvalues, and some are even lower than 50\%, which will hinder the fitting ability and expressive power of polynomial filters.

% We can see that the eigenvalues of almost all datasets have high multiplicity, and some even exceed 50\%, which seriously hinders the fitting ability and expressive power of polynomial filters.

\section{Proposed Method}

\subsection{Preliminaries}
 
Assume that we have a graph $\mathcal{G}=(\mathcal{V}, \mathcal{E},\mathbf{X})$, where $\mathcal{V}=\left\{v_1, \ldots, v_n\right\}$ denotes the vertex set of $n$ nodes, $\mathcal{E}$ is the edge set and $\mathbf{X} \in \mathbb{R}^{n \times d}$ is node feature matrix. The corresponding adjacency matrix is $\mathbf{A} \in \{0,1\}^{n \times n}$, where $\mathbf{A}_{ij}=1$ if there is an edge between nodes $v_i$ and $v_j$, and $\mathbf{A}_{ij} =0$ otherwise. The degree matrix $\mathbf{D} = \operatorname{diag}(d_1,...,d_n)$ of $\mathbf{A}$ is a diagonal matrix with its $i$-th diagonal entry as $d_i=\sum_j \mathbf{A}_{i j}$. The normalized adjacency matrix is $\mathbf{\hat A}= \mathbf{D}^{-\frac{1}{2}}\mathbf{A}\mathbf{D}^{-\frac{1}{2}}$. Let $\mathbf{I}$ denote the identity matrix. The normalized Laplacian matrix $\mathbf{\hat L}=\mathbf{I} - \mathbf{\hat A}$. Let $\mathbf{\hat L} = \mathbf{U} \boldsymbol{\Lambda} \mathbf{U}^{\top}$ denote the eigen-decomposition of $\mathbf{\hat L}$, where $\mathbf{U}$ is the matrix of eigenvectors and $\boldsymbol{\Lambda} = \text{diag} (\boldsymbol{\lambda}) = \text{diag} ( [\lambda_{1},\lambda_{2},\dots,\lambda_{n}])$ is the diagonal matrix of eigenvalues.
% Note that $\hat L = I - D^{-\frac{1}{2}} A D^{-\frac{1}{2}}$ shares the same eigenvectors with $\hat A=D^{-\frac{1}{2}} A D^{-\frac{1}{2}}$ and the eigenvalues of $\hat L$ are $\mu_i = 1 - \lambda_i$.

The Fourier transform of a graph signal $\mathbf{x}$ is defined as
$\hat{\mathbf{x}}=\mathbf{U}^{\top} \mathbf{x}$, and its inverse is $\mathbf{x}=\mathbf{U} \hat{\mathbf{x}}$. Thus the graph propagation for signal $\mathbf{x}$ with kernel $\mathbf{g}$ can be defined as:

\begin{equation}
\label{eq:Fourier_transform}
\mathbf{z}=\mathbf{g} *_{\mathcal{G}} \mathbf{x}=\mathbf{U}\left(\left(\mathbf{U}^{\top} \mathbf{g}\right) \odot \mathbf{U}^{\top} \mathbf{x}\right)=\mathbf{U} \mathbf{\hat G} \mathbf{U}^{\top} \mathbf{x} ,
\end{equation}
where $\mathbf{\hat G}=\operatorname{diag}\left(\hat g_1, \ldots,\hat  g_n\right)$ denotes the spectral kernel coefficients. In Eq. (\ref{eq:Fourier_transform}), the signal $\mathbf{x}$ is first transformed from the spatial domain to the spectral domain, then signals of different frequencies are enhanced or weakened by the filter $\mathbf{\hat G}$, and finally transformed back to the spatial domain to obtain the filtered signal $\mathbf{z}$. To avoid eigen-decomposition, current works with spectral convolution usually use the polynomial functions $h(\mathbf{\hat L})$ to approximate different kernels, we call them \textit{polynomial spectral GNNs}:
\begin{equation}
\mathbf{Z}=h(\mathbf{\hat L}) \mathbf{X} \mathbf{W}=\mathbf{U} h(\boldsymbol{\Lambda}) \mathbf{U}^{\top}  \mathbf{X} \mathbf{W}
\label{eq:filter} ,
\end{equation}
where $\mathbf{W}$ is a learnable weight matrix for feature mapping and $\mathbf{Z} $ is the prediction matrix.
% \mathbf{z}=g(\mathbf{\hat L}) \mathbf{x}=\mathbf{U} g(\boldsymbol{\Lambda}) \mathbf{U}^{\top} \mathbf{x}
% where $\hat{\Lambda}=diag(g(\lambda_1), \ldots, g(\lambda_n))$  and we call $g$ the graph filter function for above graph propagation Eq. \ref{eq:filter}.

\subsection{Expressive Power of Existing Polynomial Spectral GNNs}
% In order to focus on the fitting ability of the filter h(L), in this subsection we assume that the new feature matrix is X’=XW, and all elements of X’ are non-zero.
% For the expressive power of polynomial spectral GNNs, we have the following lemma :

For the expressive power of polynomial spectral GNNs, The following lemma is given in JacobiConv:

\begin{lemma}
\label{lemma}
polynomial spectral GNNs can produce any one-dimensional prediction if  $\mathbf{\hat L}$ has no repeated eigenvalues and $\mathbf{X}$ contains all frequency components.
\end{lemma}

The proof of this lemma can be found in Appendix B.1 of JacobiConv \cite{JacobiConv}. The lemma shows that the expressive power of polynomial spectral GNNs is related to the repeated eigenvalues.

To focus on exploring the fitting ability of the filter $h(\boldsymbol{\Lambda})$ and avoid the influence of $\mathbf{W}$, we fix the weight matrix $\mathbf{W}$ and assume that all elements of $(\mathbf{U}^{\top}\mathbf{X}\mathbf{W})$ are non-zero. Thus, for the relationship between the expressive power of polynomial spectral GNNs and the distinct eigenvalues, we have the following theorem:
\begin{theorem}
\label{th:k_eigenvalues}
When there are only $k$ distinct eigenvalues of the normalized Laplacian matrix, polynomial spectral GNNs can produce at most $k$ different filter coefficients, and thus can only generate one-dimensional predictions with a maximum of $k$ arbitrary elements.
\end{theorem}

% Theorem 3.2 rigorously states that the greater the number of distinct eigenvalues, the more powerful the expressive power of the polynomial spectral GNNs.

We provide the proof of this theorem in Appendix A.1. Theorem \ref{th:k_eigenvalues} strictly shows that when the eigenvalues are more distinct, the polynomial spectral GNNs will be more expressive. In other words, the more repeated eigenvalues, the less expressive polynomial spectral GNNs are. Thus, the number of distinct eigenvalues is the key to the expressive power of polynomial spectral GNNs. Therefore, we need to reduce the repeated eigenvalues of the normalized Laplacian matrix as much as possible, given that real-world graphs often have many repeated eigenvalues. 

\subsection{Eigenvalue Correction}
\label{subsection:Eigenvalue_Correction}
We now elaborate on the proposed eigenvalue correction strategy. In order to reduce repeated eigenvalues, a natural solution is that we can sample equally spaced from [0, 2] as new eigenvalues $\boldsymbol{\upsilon}$. Thus, the $i$-th value of $\boldsymbol{\upsilon}$ is defined as follows:
\begin{equation}
\label{eq:upsilon_i}
\upsilon_i=\frac{2i}{n-1},\quad i \in \{0,1,2,...,n-1\},\quad
\end{equation}
where $n$ is the number of nodes, which is also the number of eigenvalues. We serialize the vector $\boldsymbol{\upsilon}$ into $\{\upsilon_i\}$. Obviously, the sequence $\{\upsilon_i\}$ is a strictly monotonically increasing arithmetic sequence with 0 as the first item and $\frac{2}{n-1} $ as the common difference. We call $\boldsymbol{\upsilon}$ equidistant eigenvalues.

\begin{figure}[t] 
	\centering
	\subfigure[original eigenvalues]{\includegraphics[width=0.49\linewidth]{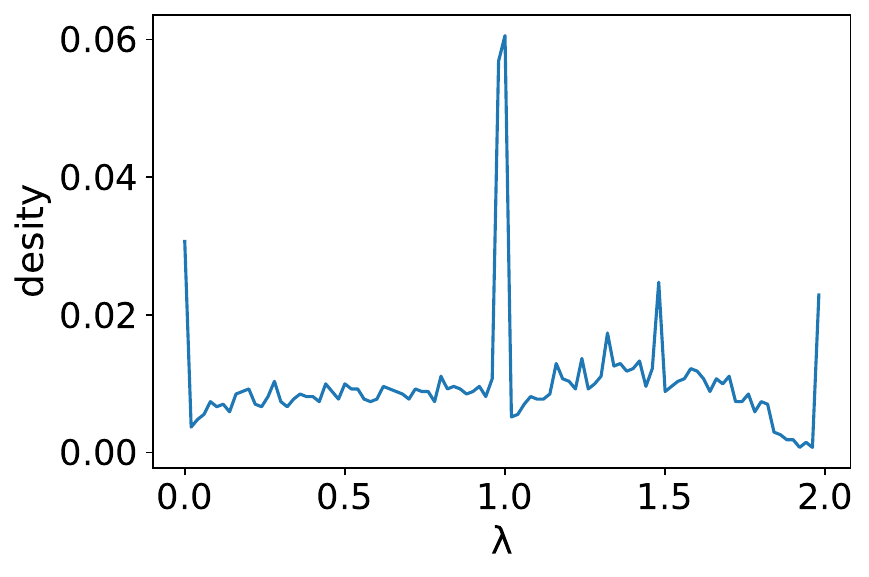}}
	\subfigure[corrected eigenvalues]{\includegraphics[width=0.49\linewidth]{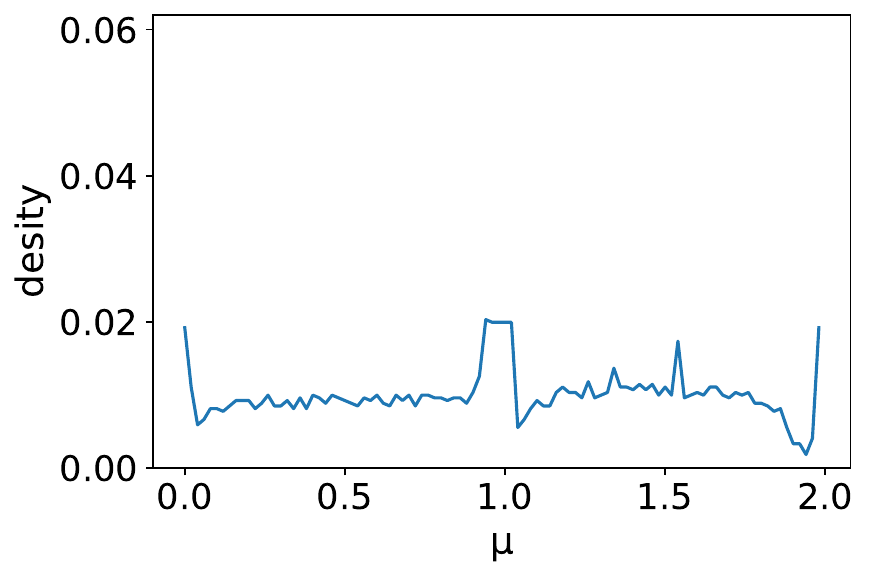}}
      % \subfigure[cora\_e\_new\_e]{\includegraphics[width=0.6\linewidth]{images/cora_e_new_e.png}}
	\caption{Histogram of the probability density distribution of the original eigenvalues and the corrected eigenvalues.}
	\label{fig:lambda_mu}
 % \vspace{-3mm}
\end{figure}

However, the original eigenvalue $\boldsymbol{\lambda}$ of the normalized Laplacian matrix describes the frequency information and represents the speed of the signal change. Specifically, the magnitude of the eigenvalue represents the difference information of the signal between the neighbors of the node. Therefore, if we directly use the $\boldsymbol{\upsilon}$ in Eq. (\ref{eq:upsilon_i}) as the new eigenvalues, polynomial spectral GNNs will not be able to capture the frequency information represented by each frequency component. Hence, it is still necessary to include the original eigenvalues in the input. A simple way is to amalgamate them linearly to obtain the new eigenvalues $\boldsymbol{\mu}$:
\begin{equation}
\mu_i= \beta \lambda_i +(1-\beta)\upsilon_i ,\quad i \in \{0,1,2,...,n-1\},\quad
\label{eq:beta_i_ei}
\end{equation}
where $\beta \in [0,1] $ is a tunable hyperparameter, and $\{\lambda_i\}$ is a sequence of original eigenvalues $\boldsymbol{\lambda}$ in ascending order. Thus, we have the following theorem:
\begin{theorem}
\label{monotonically_increasing}
When $\beta \in [0,1) $, the sequence $\{\mu_i\}$ is strictly monotonically increasing, that is, all n elements of the sequence $\{\mu_i\}$ are different from each other.
\end{theorem}

The proof of this theorem is given in Appendix A.2. Theorem \ref{monotonically_increasing} indicates that the new eigenvalues $\boldsymbol{\mu} = \{\mu_1,...,\mu_n \}$ have no repeated elements when $\beta \in [0,1)$, thus satisfying the condition of Lemma \ref{lemma}. It should be noted that if the spacing between two eigenvalues is very small, it is difficult for a polynomial spectral GNN to generate different filter coefficients for them. Therefore, the spacing between eigenvalues is crucial to guarantee the expressive power of polynomial spectral GNNs.

It is worth noting that $\beta$ is a key hyperparameter in Eq. (\ref{eq:beta_i_ei}), which can be used to trade off between $\{\lambda_i\}$ and $\{\upsilon_i\}$. The smaller $\beta$, the closer $\{\mu_i\}$ is to $\{\upsilon_i\}$, and the more uniform the spacing between the new eigenvalues $\{\mu_i\}$, but the frequency information may not be captured. On the contrary, the larger $\beta$, the closer $\{\mu_i\}$ is to $\{\lambda_i\}$, and the closer the new eigenvalue is to the frequency it represents, but the spacing between the eigenvalues may be uneven. When $\beta = 1$ , $\{\mu_i\}$ degenerates into the original eigenvalues $\{\lambda_i\}$ of the Laplacian matrix. When $\beta = 0$, $\{\mu_i\}$ degenerates to the equidistantly sampled $\{\upsilon_i\}$ in Eq. (\ref{eq:upsilon_i}), whose spacing is absolutely uniform. In short, the hyperparameter $\beta$ needs to be set to an appropriate value so that the spacing between eigenvalues is relatively uniform and not too far away from the original eigenvalues.

For example, the Figure \ref{fig:lambda_mu} shows the distribution histogram of the original eigenvalues $\{\lambda_i\}$ and the corrected eigenvalues $\{\mu_i\}$ on Cora dataset when $\beta$=0.5. We can see that, compared to Figure \ref{fig:lambda_mu}(a), Figure \ref{fig:lambda_mu}(b) has no sharp bumps, and the distribution of eigenvalues is more uniform. Hence, it is expected to fit more complex and diverse filters.

\subsection{Combination of Corrected Eigenvalues with Existing Polynomial Filters}
In order to fully take advantage of polynomial spectral GNNs, after getting the corrected eigenvalues, We are required to integrate them with a polynomial filter. First, we set $\mathbf{E}=\mathbf{U}\operatorname{diag}(\boldsymbol{\upsilon})\mathbf{U}^{\top}$, $\mathbf{H}=\mathbf{U}\operatorname{diag} (\boldsymbol{\mu}) \mathbf{U}^{\top}$. Combined with Eq. (\ref{eq:beta_i_ei}) and $\mathbf{\hat L} = \mathbf{U} \operatorname{diag}(\boldsymbol{\lambda}) \mathbf{U}^{\top}$, we have:
\begin{equation}
\mathbf{H}=\beta \mathbf{\hat L} + (1-\beta) \mathbf{E} . 
\label{eq:H_L}
\end{equation}
Then, a polynomial $h(\mathbf{H})$ of order $K$ can be written as:
\begin{equation}
\label{eq:polynomials_matrix}
h(\mathbf{H}) = \sum_{k=0}^{K}\alpha_k \mathbf{H}^k = \sum_{k=0}^{K}\alpha_k (\beta \mathbf{\hat L}+(1-\beta) \mathbf{E})^k ,
\end{equation}
where $h(x)=\sum_{k=0}^K \alpha_k x^k$. The difference between the proposed method and the existing polynomial filter is that $\mathbf{\hat L}$ is replaced by $\mathbf{H}$. Therefore, our method can be quite flexibly integrated into existing polynomial filters without excessive modifications. However, when the degree of the polynomial is high, the matrix multiplication will be computationally expensive, especially for dense graphs. Fortunately, we can compute $h(\mathbf{H})$ by directly using polynomials of the eigenvalues instead of polynomials of the matrix:
\begin{equation}
\label{eq:polynomials_eigenvalues}
h(\mathbf{H}) = h(\mathbf{U}\operatorname{diag}(\boldsymbol{\mu})\mathbf{U}^{\top})  = \mathbf{U} \sum_{k=0}^{K}\operatorname{diag}(\alpha_k \boldsymbol{\mu}^k) \mathbf{U}^{\top} .
\end{equation}
We find that using Eq. (\ref{eq:polynomials_eigenvalues}) to calculate $h(\mathbf{H})$ is much faster than Eq. (\ref{eq:polynomials_matrix}), resulting in improved training efficiency. Moreover, we can increase the order $K$ to improve the fitting ability without increasing too much computational cost.

The complete filtering process can be expressed as:
\begin{equation}
\mathbf{Z}=h(\mathbf{H}) \mathbf{X} \mathbf{W}=\mathbf{U} \sum_{k=0}^{K}\operatorname{diag}(\alpha_k \boldsymbol{\mu}^k) \mathbf{U}^{\top} \mathbf{X} \mathbf{W} .
\end{equation}
Next, we present the details of the proposed eigenvalue correction strategy integrated into existing polynomial filters.

\subsubsection{Insertion into GPR-GNN.}
GPR-GNN \cite{GPR-GNN} directly assigns a learnable coefficient to each order of the normalized adjacency matrix $\mathbf{\hat A}$, and its polynomial filter is defined as:
\begin{equation}
\sum_{k=0}^K \gamma_k  \mathbf{\hat A}^k = \mathbf{U} g_{\gamma, K}(\Lambda) \mathbf{U}^T ,
\end{equation}
where $g_{\gamma, K}(\Lambda)$ is an element-wise operation, and $g_{\gamma, K}(x)=\sum_{k=0}^K \gamma_k x^k$. It is worth noting that GPR-GNN uses a polynomial of the normalized adjacency matrix instead of the Laplacian matrix. Fortunately, they are constrained via : $\mathbf{\hat A}  =\mathbf{I}-\mathbf{\hat L}$. Therefore, we integrate the corrected eigenvalues $\boldsymbol{\mu}$ into GPR-GNN to obtain a new polynomial filter:
\begin{equation}
\sum_{k=0}^K \gamma_k  \mathbf{\hat A}^k =\mathbf{U} \sum_{k=0}^K \operatorname{diag}( \gamma_k (1-\boldsymbol{\mu})^k )\mathbf{U}^T .
\end{equation}

\subsubsection{Insertion into BernNet.}
BernNet \cite{BernNet} expresses the filtering operation with Bernstein polynomials and forces all coefficients to be positive, and its filter is defined as:
\begin{equation}
\sum_{k=0}^K \theta_k \frac{1}{2^K}\left(\begin{array}{c}
K \\
k
\end{array}\right)(2 \mathbf{I}-\mathbf{L})^{K-k} \mathbf{L}^k .
\end{equation}
Similarly, we have:
\begin{equation}
\mathbf{U}  \sum_{k=0}^K  \operatorname{diag}( \theta_k \frac{1}{2^K}\left(\begin{array}{l} 
K \\
k
\end{array}\right)(2-\boldsymbol{\mu})^{K-k} \odot \boldsymbol{\mu}^k )\mathbf{U}^{\top} .
\end{equation}
\subsubsection{Insertion into JacobiConv.}
JacobiConv \cite{JacobiConv} proposes a Jacobi basis to adapt a wide range of weight functions due to its orthogonality and flexibility. The iterative process of Jacobi basis can be defined as:
\begin{equation}
\begin{aligned}
& P_0^{a, b}(x)=1 , \\
& P_1^{a, b}(x)=0.5 a-0.5 b+(0.5 a+0.5 b+1) x , \\
& P_k^{a, b}(x)=(2 k+a+b-1) \\
& \cdot \frac{(2 k+a+b)(2 k+a+b-2) x +a^2-b^2}{2 k(k+a+b)(2 k+a+b-2)} P_{k-1}^{a, b}(x) \\
& -\frac{(k+a-1)(k+b-1)(2 k+a+b)}{k(k+a+b)(2 k+a+b-2)} P_{k-2}^{a, b}(x) ,
\end{aligned}
\end{equation}
where $a$ and $b$ are tunable hyperparameters. Unlike GPR-GNN and BernNet, JacobiConv adopts an individual filter function for each output dimension $l$:
\begin{equation}
\mathbf{Z}_{: l}=\sum_{k=0}^K \alpha_{k l} P_k^{a, b}(\mathbf{\hat A}) (\mathbf{X}\mathbf{W})_{: l} .
\end{equation}
Then, we integrate the corrected eigenvalues $\boldsymbol{\mu}$ into the JacobiConv:
\begin{equation}
\mathbf{Z}_{: l}=\mathbf{U} \sum_{k=0}^K \operatorname{diag}( \alpha_{k l} \cdot P_k^{a, b}(1-\boldsymbol{\mu}) )\mathbf{U}^T (\mathbf{X}\mathbf{W})_{: l} .
\end{equation}
% Note that JacobiConv employs a Polynomial Coefficient Factorization (PCD) technique, which is not presented here due to space constraints.

\section{Experiment}

% \subsection{Settings}
In this section, we conduct a series of comprehensive experiments to demonstrate the effectiveness of the proposed method. First, we integrate the eigenvalue correction strategy into the three methods of GPR-GNN, BernNet, and JacobiConv to complete the learning of filter functions on synthetic datasets and node classification tasks on real-world datasets. Then, we report experiments on ablation analysis and sensitivity analysis of the hyperparameter $\beta$. Finally, we show the evaluation of the time cost. All experiments are conducted on a machine with 3 NVIDIA A5000 24GB GPUs and Intel(R) Xeon(R) Silver 4310 2.10GHz CPU. 
% The code is available in the supplementary material.
\subsection{Evaluation on Learning Filters}
\label{subsection:Learning_Filters}
Following previous works \cite{BernNet,JacobiConv}, we transform 50 real images to 2D regular 4-neighbor grid graphs. The nodes are pixels, and every two neighboring pixels have an edge between the corresponding nodes. Five predefined graph filters, i.e., low ($e^{-10\lambda^2}$), high ($1-e^{-10\lambda^2}$), band ($e^{-10(\lambda-1)^2}$), reject ($1-e^{-10(\lambda-1)^2}$), and comb ($|\sin \pi\lambda|$) are used to generate ground truth graph signals. We use the mean squared error of 50 images as the evaluation metric.

To verify the effectiveness of the proposed \textbf{E}igenvalue \textbf{C}orrection (EC) strategy, we integrate it into three representative spectral GNNs with trainable polynomial filters, i.e., GPR-GNN \cite{GPR-GNN}, BernNet \cite{BernNet}, and JacobiConv \cite{JacobiConv}. Thus, we have three variants: EC-GPR, EC-Bern, and EC-Jacobi. For fairness, all experimental settings are the same as the base model, including dataset partition, hyperparameters, random seeds, etc. We just tune the $\beta \in \{0,0.1,...,0.9\} $ hyperparameter to select the best model. In addition, for a more comprehensive comparison, we also consider four other baseline methods, including GCN \cite{GCN}, GAT \cite{GAT}, ARMA \cite{ARMA}, and ChebNet \cite{Chebyshev}. 

The quantitative experiment results are shown in Table \ref{table:filter}, revealing that our methods all achieve the best performance compared to the base models. For example, on low-pass and high-pass filters, our method outperforms the base model by an average of 12.9\%, while on band-pass, comb-pass, and reject-pass filters, our method outperforms the base model by an average of 77.9\%. Since complex filters require more frequency components to fit, our method improves more on band-pass, comb-pass, and reject-pass filters. Moreover, GCN and GAT only perform better on low-pass filters applied to homophilic graphs, illustrating the limited fitting ability of fixed filters. In contrast, polynomial-based GNNs perform better due to the ability to learn arbitrary filters. Nevertheless, they all take a large number of repeated eigenvalues as input, therefore, their expressive power is still weaker than our methods.

\begin{table}[t]
\centering
%\vskip -0.05in
% \vskip 0.1in
\begin{center}
\begin{small}
\setlength{\tabcolsep}{1mm}
{\begin{tabular}{lccccc}
\hline
           & Low      & High     & Band     & Reject & Comb    \\ 
           \hline
GCN       &$3.4799$ &$67.6635$ &$25.8755$ &$21.0747$ &$50.5120$  \\
GAT       &$2.3574$ &$21.9618$ &$14.4326$ & $12.6384$ &$23.1813$  \\
ARMA       & $1.8478$ & $1.8632$ & $7.6922$ & $8.2732$ & $15.1214$ \\
ChebyNet   & $0.8220$ & $0.7867$ & $2.2722$ & $2.5296$  & $4.0735$  \\
\hline
GPR-GNN    & $0.4169$ & $0.0943$ & $3.5121$ & $3.7917$  & $4.6549$ \\
\bf{EC-GPR}    & $\bf{0.2703}$ & $\bf{0.0656}$ & $\bf{0.2920}$ & $\bf{1.1266}$  & $\bf{1.0681}$ \\
\hline
BernNet    & $0.0314$ & $\bf{0.0113}$ & $0.0411$ & $0.9313$ & $0.9982$  \\
\bf{EC-Bern}   & $\bf{0.0277}$ & $\bf{0.0113}$ & $\bf{0.0058}$ & $\bf{0.2391}$ & $\bf{0.3302}$ \\
\hline
JacobiConv	&$\bf{0.0003}$  &$\bf{0.0011}$  &$0.0213$  &$0.0156$  &$0.2933$\\
\bf{EC-Jacobi}	&$\bf{0.0003}$	&$\bf{0.0011}$	&$\bf{0.0088}$	&$\bf{0.0018}$ &$\bf{0.0370}$\\
\hline
\end{tabular}}
\end{small}
\end{center}
\caption{Average loss in learning filters experiments.}
\label{table:filter}
% \vskip -0.05in
\end{table}

\begin{table*}[t]
\centering
% \vskip 0.1in
\begin{center}
\begin{small}
\resizebox{\textwidth}{!}{
\setlength{\tabcolsep}{1mm}

\begin{tabular}{lccc|cc|cc|cc}
\hline
% \toprule
Datasets  & GCN        & APPNP      & Chebynet   & GPR-GNN     & \bf{EC-GPR}     & BernNet    & \bf{EC-Bern} & JacobiConv          & \bf{EC-Jacobi} \\ 
\hline
% \midrule
Cora  & $87.14_{\pm1.01}$ & $88.14_{\pm0.73}$ & $86.67_{\pm0.82}$ & $88.57_{\pm0.69}$ & $\bf{89.38_{\pm0.95}}$ & $88.52_{\pm0.95}$ & $\bf{88.65_{\pm0.87}}$ & $88.98_{\pm0.46}$ & $\bf{89.00_{\pm0.67}}$    \\
Citeseer & $79.86_{\pm0.67}$ & $80.47_{\pm0.74}$ & $79.11_{\pm0.75}$ & $80.12_{\pm0.83}$ & $\bf{80.66_{\pm0.85}}$ & $80.09_{\pm0.79}$ & $\bf{80.26_{\pm0.60}}$ & $80.78_{\pm0.79}$ & $\bf{81.23_{\pm1.00}}$    \\
Pubmed  & $86.74_{\pm0.27}$ & $88.12_{\pm0.31}$ & $87.95_{\pm0.28}$ & $88.46_{\pm0.33}$ & $\bf{89.63_{\pm0.31}}$ & $88.48_{\pm0.41}$ & $\bf{89.10_{\pm0.34}}$ & $89.62_{\pm0.41}$ & $\bf{89.64_{\pm0.37}}$    \\
Computers  & $83.32_{\pm0.33}$ & $85.32_{\pm0.37}$ & $87.54_{\pm0.43}$ & $86.85_{\pm0.25}$ & $\bf{89.89_{\pm0.43}}$ & $87.64_{\pm0.44}$ & $\bf{88.36_{\pm0.48}}$ & $\bf{90.39_{\pm0.29}}$ & $90.33_{\pm0.28}$    \\
Photo  & $88.26_{\pm0.73}$ & $88.51_{\pm0.31}$ & $93.77_{\pm0.32}$ & $93.85_{\pm0.28}$ & $\bf{94.76_{\pm0.94}}$ & $93.63_{\pm0.35}$ & $\bf{94.49_{\pm0.30}}$ & $95.43_{\pm0.23}$ & $\bf{95.54_{\pm0.34}}$   \\
Chameleon  & $59.61_{\pm2.21}$ & $51.84_{\pm1.82} $ & $59.28_{\pm1.25}$ & $67.28_{\pm1.09}$ & $\bf{74.20_{\pm1.07}}$ & $68.29_{\pm1.58}$ & $\bf{74.20_{\pm1.31}}$ & $74.20_{\pm1.03}$ & $\bf{75.62_{\pm1.51}}$   \\
Actor   & $33.23_{\pm1.16}$ & $39.66_{\pm0.55}$ & $37.61_{\pm0.89}$ & $39.92_{\pm0.67}$ & $\bf{40.43_{\pm0.84}}$ & $41.79_{\pm1.01}$ & $\bf{41.88_{\pm1.05}}$ & $\bf{41.17_{\pm0.64}}$ & $40.99_{\pm0.70}$ \\
Squirrel & $46.78_{\pm0.87}$ & $34.71_{\pm0.57}$ & $40.55_{\pm0.42}$ & $50.15_{\pm1.92}$ & $\bf{62.43_{\pm0.67}}$ & $51.35_{\pm0.73}$ & $\bf{62.74_{\pm1.04}}$  & $57.38_{\pm1.25}$ & $\bf{59.88_{\pm0.86}}$ \\
Texas  & $77.38_{\pm3.28}$ & $90.98_{\pm1.64}$ & $86.22_{\pm2.45}$ & $\bf{92.95_{\pm1.31}}$ & $92.30_{\pm1.80}$  & $93.12_{\pm0.65}$ &$\bf{94.43_{\pm1.31}}$  & $93.44_{\pm2.13}$ & $\bf{93.44_{\pm1.48}}$  \\
Cornell & $65.90_{\pm4.43}$ & $91.81_{\pm1.96}$ & $83.93_{\pm2.13}$ & $\bf{91.37_{\pm1.81}}$ & $90.66_{\pm2.13}$ & $92.13_{\pm1.64}$ & $\bf{93.77_{\pm0.98}}$ & $92.95_{\pm2.46}$ & $\bf{93.28_{\pm2.30}}$\\
\hline
% \bottomrule
\end{tabular}
}
\end{small}
\end{center}
\caption{Results on real-world datasets: Mean accuracy (\%) $\pm$ $95$\% confidence interval.}
\label{tab:real}
\end{table*}

\subsection{Evaluation on Real-World Datasets}
We also evaluate the proposed method on the node classification task on real-world datasets. Following JacobiConv \cite{JacobiConv}, for the homophilic graphs, we employ three citation datasets Cora, CiteSeer and PubMed \cite{citation_dataset}, and two co-purchase datasets Computers and Photo \cite{co-purchase_dataset}. For heterophilic graphs, we use Wikipedia graphs Chameleon and Squirrel \cite{Wikipedia_dataset}, the Actor co-occurrence graph, and the webpage graph Texas and Cornell from WebKB3 \cite{webpage_dataset}. Their statistics are listed in Table \ref{tab:datasets}.

Following previous work \cite{JacobiConv}, we use the full-supervised split, i.e., 60\% for training, 20\% for validation, and 20\% for testing. Mean accuracy and 95\% confidence intervals over 10 runs are reported. We integrate the eigenvalue correction strategy into GPR-GNN \cite{GPR-GNN}, BernNet \cite{BernNet}, and JacobiConv \cite{JacobiConv}. Moreover, we also add the results of GCN \cite{GCN}, APPNP \cite{PPNP}, and ChebyNet \cite{Chebyshev}. For a fair comparison, we only tune the hyperparameter $\beta \in \{0,0.01,...,0.99\}$, and the other experimental settings remain the same as the base model. All hyperparameter $\beta$ settings can be found in Appendix D.

\begin{table*}[t]
\centering
% \vskip 0.1in
\begin{center}
\begin{small}
\resizebox{0.9\textwidth}{!}{
\setlength{\tabcolsep}{2.5mm}
\begin{tabular}{l|cc|cc|cc|c}
\hline Datasets  & GPR-GNN    & EC-GPR    & BernNet    & EC-Bern   & JacobiConv & EC-Jacobi & Decomposition  \\
\hline
Cora      & 6.73/1.40 & 4.09/0.84 & 22.25/6.06 & 4.73/1.28 & 9.50/5.02  & 9.29/3.74   &  0.62\\
Citeseer  & 6.90/1.46 & 4.09/0.93 & 22.12/6.46 & 4.39/1.29 & 9.39/4.71  & 9.45/3.07   &  0.82\\
Pubmed    & 7.18/1.46 & 9.15/1.86 & 20.45/8.70 & 9.73/4.15 & 9.90/7.63  & 13.79/11.26 &  58.59\\
Computers & 8.20/1.88 & 7.83/1.09 & 29.40/8.32 & 8.49/2.31 & 9.81/5.58  & 11.91/7.92  &  27.53\\
Photo     & 7.09/1.56 & 4.54/1.23 & 21.04/8.74 & 5.26/2.21 & 9.23/6.66  & 9.97/8.18   &  3.86\\
Chameleon & 7.31/1.65 & 4.12/1.00 & 22.61/5.12 & 5.01/1.18 & 9.32/7.47  & 9.62/7.10   &  0.27\\
Actor     & 7.17/2.54 & 4.53/1.49 & 18.74/6.68 & 5.44/2.02 & 9.37/4.25  & 9.97/3.86   &  3.56\\
Squirrel  & 7.23/3.65 & 3.96/1.75 & 22.66/7.54 & 2.42/0.63 & 9.44/8.16  & 9.60/7.86   &  1.61\\
Texas     & 6.83/1.41 & 3.80/0.79 & 22.42/4.76 & 4.50/0.95 & 9.58/6.47  & 9.33/6.48   &  0.03\\
Cornell   & 7.17/1.48 & 3.89/0.80 & 22.34/4.66 & 4.49/0.93 & 9.10/5.30  & 9.53/3.18   &  0.02\\
\hline
\end{tabular}
}
\end{small}
\end{center}
\caption{Per-epoch time (ms)/total training time (s) and eigen-decomposition time (s).}
\label{tab:time}
\end{table*}

Table \ref{tab:real} presents the experimental results on real-world datasets, from which we have the following observations. 1) GNNs with learnable filters tend to perform better than GNNs with fixed filters since fixed filters, such as GCN and APPNP, cannot learn complex filter responses. In contrast, GPR-GNN, BernNet, and JacobiConv can learn different types of filters from both homophilic graphs and heterophilic graphs. 2) The proposed method outperforms the base models in at least eight of the ten datasets, especially on the Squirrel dataset, with an average improvement of 17\% over the three base models. This is because heterophilic graphs require more frequency components than homophilic graphs to fit the patterns. In general, we achieve a 4.2\% improvement in GPR-GNN, 3.7\% improvement in BernNet, and 0.68\% improvement in JacobiConv, demonstrating that our proposed eigenvalue correction strategy can improve the expressive power of polynomial spectral GNNs. 3) With our eigenvalue correction strategy, GPR-GNN and BernNet achieve competitive performance with JacobiConv, which shows that the expressive power of different polynomial filters with the same order is the same.

\subsection{Ablation Analysis}
\label{sec:abl}
\begin{figure}[t]   
	\centering 
	\includegraphics[width=\linewidth]{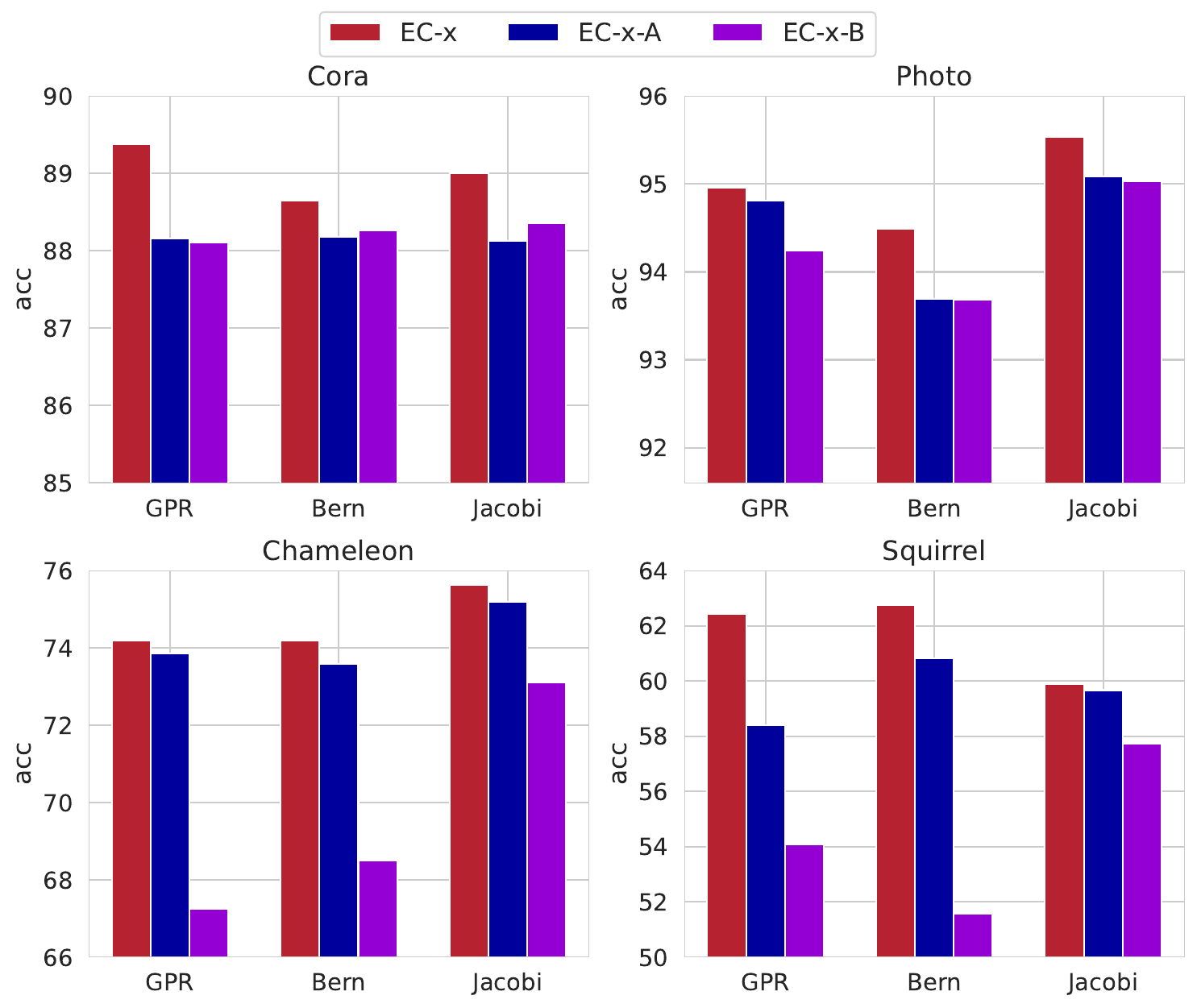}
	\caption{Ablation study of proposed method EC-$x$ on four datasets with our
variants EC-$x$-A and EC-$x$-B for all $x \in$ \{GPR, Bern, Jacobi\}. }
 % \vspace{-4mm}
\end{figure}
This subsection aims to validate our design through ablation studies. Previously, we believe that neither the original eigenvalues nor the equally spaced eigenvalues are optimal, while only combining them in a certain proportion can achieve the optimal performance. Therefore, we design two variants EC-$x$-A and EC-$x$-B to verify our conjecture, where $x \in $\{GPR, Bern, Jacobi\}. EC-$x$-A directly uses the equidistant eigenvalues $\boldsymbol{\upsilon}$ as input, that is, $\beta$ in Eq. (\ref{eq:beta_i_ei}) equals to 0. In contrast, EC-$x$-B directly uses the original eigenvalues $\boldsymbol{\lambda}$ as input, that is, $\beta$ in Eq. (\ref{eq:beta_i_ei}) equals to 1.

Figure 4 shows the results of the ablation experiments on two homophilic graphs, Cora and Photo, and two heterophilic graphs, Chameleon and Squirrel. From Figure 4, we have the following findings: 1) The utilization of solely the original eigenvalues and equidistant eigenvalues as input demonstrates a decrease in performance, corroborating our hypothesis necessitating their amalgamation. 2) EC-$x$-B of the heterophilic graphs drops more than the homophilic graphs, which indicates that the heterophilic graphs need more frequency component information to fit various responses.

\subsection{Parameter Sensitivity Analysis}
% In subsection \ref{subsection:Eigenvalue_Correction}
As discussed above, we add a key hyperparameter $\beta$ to adjust the ratio of the original eigenvalue $\boldsymbol{\lambda}$ and the equidistant eigenvalue $\boldsymbol{\upsilon}$. Now we analyze the impact of this hyperparameter on the performance of homophilic graphs and heterophilic graphs. Similar to the ablation experiments, we select two homophilic datasets Cora and Photo, and two heterophilic datasets Chameleon and Squirrel. Figure 5 shows the effect of the hyperparameter $\beta$ on the performance of the three methods EC-GPR, EC-Bern, and EC-Jacobi.

Figure 5 indicates that on the Cora dataset, the general trend of model performance is to decrease with the increase of $\beta$, and then increase, which shows that both the equidistant eigenvalues and original eigenvalues are useful. The trend on the Photo dataset is not prominent, possibly due to its limited occurrence of repeated eigenvalues. In comparison, for heterophilic graphs, the model performance is maintained at a high level when $\beta$ is less than 0.6, and when $\beta$ is greater than 0.6, the model performance drops significantly. This suggests that equidistant eigenvalues contribute more to heterophilic graphs, which is consistent with our previous analysis.

\begin{figure}[t]   
	\centering 
	\includegraphics[width=\linewidth]{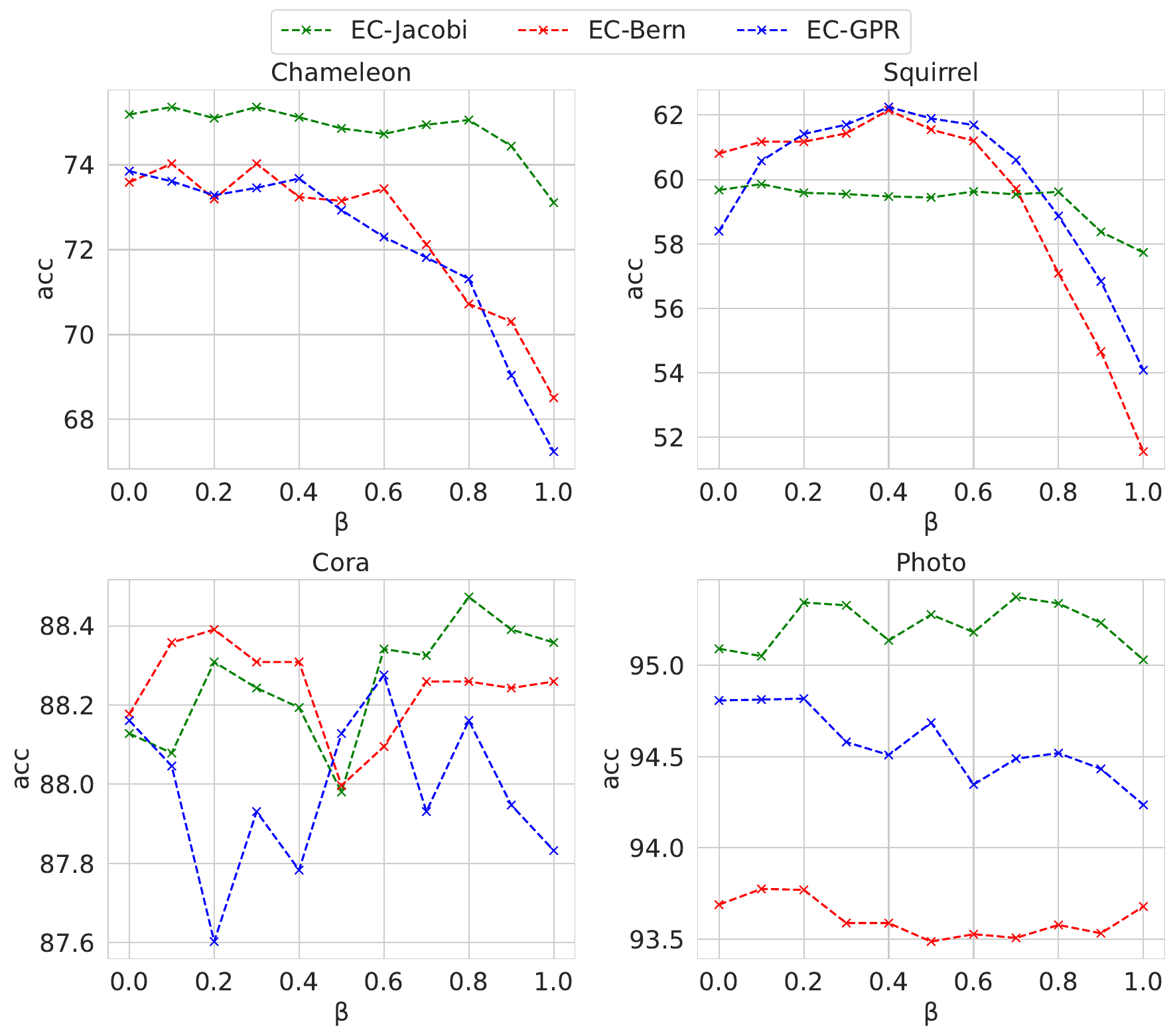}
	\caption{Effect of hyperparameter $\beta$ on model performance. }
 % \vspace{-4mm}
 \label{fig:histogram}
\end{figure}

\subsection{Evaluation on Time Cost}
Table 4 shows our time cost compared with the base models GPR-GNN, BernNet and JacobiConv. We can conclude that our time spent with the JacobiConv model is similar, 30.9\% less than GPR-GNN and 75.5\% less than BernNet. In particular, we significantly improve the training efficiency of BernNet, which is a quadratic time complexity related to the order $K$ of its polynomial. This is because we change the time-consuming polynomial of the matrix to the polynomial of the eigenvalues, thus reducing the calculation time.

Since obtaining eigenvalues and eigenvectors requires eigen-decomposition, we also add the time overhead of eigen-decomposition in Table 4. We can see from Table 4, the eigen-decomposition time for most datasets is less than 5 seconds, even for a large graph with nearly 20,000 nodes like Pubmed, the time for eigen-decomposition is less than one minute. For larger graphs, we can follow previous works \cite{SignNet,Specformer,SPAN,EigLearn} and only take the largest and smallest $k$ eigenvalues and eigenvectors without a full eigen-decomposition. Moreover, the eigen-decomposition only needs to be precomputed once and can be reused in training afterward. In summary, the cost of eigen-decomposition is acceptable and negligible.

\section{Conclusions}
In this paper, we point out that the current polynomial spectral GNNs have insufficient expressive power due to a large number of repeated eigenvalues, so we propose a novel eigenvalue correction strategy that amalgamates the original eigenvalues with equidistant eigenvalues counterparts to make the distribution of eigenvalues more uniform, and not to deviate too far from the original frequency information. Moreover, we improve the training efficiency without increasing the precomputation time too much, which allows us to efficiently increase the order of polynomial spectral GNNs. Extensive experiments on synthetic and real-world datasets demonstrate that the proposed eigenvalue correction strategy outperforms the base model.

\section{Acknowledgments}
The work was supported by the National Natural
Science Foundation of China (Grant No. U22B2019), National key research and development plan (Grant No. 2020YFB2104503), BUPT Excellent Ph.D. Students Foundation(CX2023128) and CCF-Baidu Open Fund.

\bibliography{aaai24}

\appendix
\setcounter{secnumdepth}{2} 
\section{Proofs of Theorems}
Below we show proofs of several theorems in the main text of the paper.

\subsection{Proof of Theorem 1}
\textit{Proof.} We first prove the first part of Theorem 1. We know from Eq. 2 in the paper that the filter coefficients of polynomial spectral GNNs are generated by $h(\lambda)$. We assume that the polynomial $h(\lambda)$ is of sufficiently high order so that it can produce different outputs $h(\lambda_i)$ for $k$ different inputs $\lambda_i$. However, when $\lambda_i=\lambda_j$, $h(\lambda_i)$ = $h(\lambda_j)$. That is, $h(\lambda)$ can only produce the same output for the same input. Therefore, when the normalized Laplacian matrix $\mathbf{\hat L}$ has only $k$ different eigenvalues, polynomial spectral GNNs can only produce at most $k$ different filter coefficients.

Next, we prove the second part of Theorem 1. For the filtering process $\mathbf{Z}=\mathbf{U}\operatorname{diag}(h(\boldsymbol{\lambda}))\mathbf{U}^{\top}\mathbf{X}\mathbf{W}$. We first assume all elements of $\mathbf{U}^{\top}\mathbf{X}\mathbf{W}$ are non-zero. And we  set vector $\mathbf{c}=\mathbf{U}^{\top}\mathbf{X}\mathbf{W}$ and $\boldsymbol{\theta}=h(\boldsymbol{\lambda}) \odot \mathbf{c}$ . Thus we have $\mathbf{Z}=\mathbf{U}\operatorname{diag}(h(\boldsymbol{\lambda}))\cdot\mathbf{c} = \mathbf{U} \boldsymbol{\theta}$ where $\mathbf{Z} \in \mathbb{R}^{n \times 1}$.

It can be seen that if $\lambda_i=\lambda_j$, then $\theta_j=\theta_i \frac{c_j}{c_i}$. Given that we have $k$ distinct eigenvalues, only $k$ elements of the vector $\boldsymbol{\theta}$ are arbitrary. Therefore, we can always represent Z by a vector $\boldsymbol{\theta}^{'}$ of $k$ elements and a matrix $\mathbf{V}$ of $n$ rows and $k$ columns. Thus, we have $\mathbf{V} \boldsymbol{\theta}^{'}=\mathbf{Z}$:
\begin{equation}
\left(\begin{matrix}
\mathbf{V}_{11} & \mathbf{V}_{12}  & \cdots & \mathbf{V}_{1k} \\
\mathbf{V}_{21} & \mathbf{V}_{22} &  \cdots & \mathbf{V}_{2k} \\
\vdots & \vdots  & \ddots & \vdots \\
\mathbf{V}_{n1}  & \mathbf{V}_{n2}&  \cdots & \mathbf{V}_{nk}
\end{matrix}\right)
\left(\begin{matrix}
\theta_1^{'} \\
\theta_2^{'} \\
\vdots \\
\theta_k^{'} 
\end{matrix}\right)
= 
\left(\begin{matrix}
\mathbf{Z}_1 \\
\mathbf{Z}_2 \\
\vdots \\
\mathbf{Z}_n
\end{matrix}\right),
\end{equation}
where $k \leq n$ is the number of distinct eigenvalues. We can conclude that if Eq. (1) has a solution, it has at least $n-k$ redundant equations. Without loss of generality, we assume that the last $n-k$ equations are redundant, so we have: ${\forall} j \in \{k+1,\cdots,n\}$, $\mathbf{Z}_j=\alpha_1 \mathbf{Z}_1 +\cdots+\alpha_k \mathbf{Z}_k$ where $\alpha_i$ is a constant. Therefore, given $\mathbf{Z}_1$,$\cdots$,$\mathbf{Z}_k$, we can uniquely determine $\mathbf{Z}_{k+1}$,$\cdots$,$\mathbf{Z}_n$, thus polynomial spectral GNNs can only generate one-dimensional predictions with a maximum of $k$ arbitrary elements.

\subsection{Proof of Theorem 2}
We formalize Theorem 2 in the paper as follows:
\begin{theorem}
When $\beta \in [0,1)$, ${\forall} i \in \{1,\cdots,n\}$, $\mu_i > \mu_{i-1}$.
\end{theorem}

\textit{Proof.} When $\beta \in [0,1)$, given $\lambda_{i}\ge \lambda_{i-1}$, we have $\beta \lambda_{i}\ge \beta \lambda_{i-1}$ where $i \in \{1,\cdots,n\}$.

Similarly, when $\beta \in [0,1)$, given $\upsilon_{i} > \upsilon_{i-1}$, we have $(1-\beta) \upsilon_{i} > (1-\beta) \upsilon_{i-1}$ where $i \in \{1,\cdots,n\}$.

Combining the above two conclusions, we can get the following inequality:
\begin{equation}
  \mu_{i}=\beta \lambda_{i} + (1-\beta)\upsilon_{i}>\beta \lambda_{i-1} + (1-\beta)\upsilon_{i-1}=\mu_{i-1}
\end{equation}
Therefore, when $\beta \in [0,1)$, ${\forall} i \in \{1,\cdots,n\}$, $\mu_i > \mu_{i-1}$.

\section{Eigenvalue distribution of normalized Laplacian matrix}
Figure 6 shows the distribution of the eigenvalues of the normalized Laplacian matrix in different datasets, from which we can see that the eigenvalues of all datasets are clustered around 1, so there will be many repeated eigenvalues. 

% Moreover, we find that the distribution of eigenvalues is similar for the same type of datasets.

% \footnote{The types of the two datasets within the same row are the same.}
\begin{figure}[t] 
	\centering
	\subfigure[Cora]{\includegraphics[width=0.48\linewidth]{images/cora.pdf}}
        \subfigure[CiteSeer]{\includegraphics[width=0.48\linewidth]{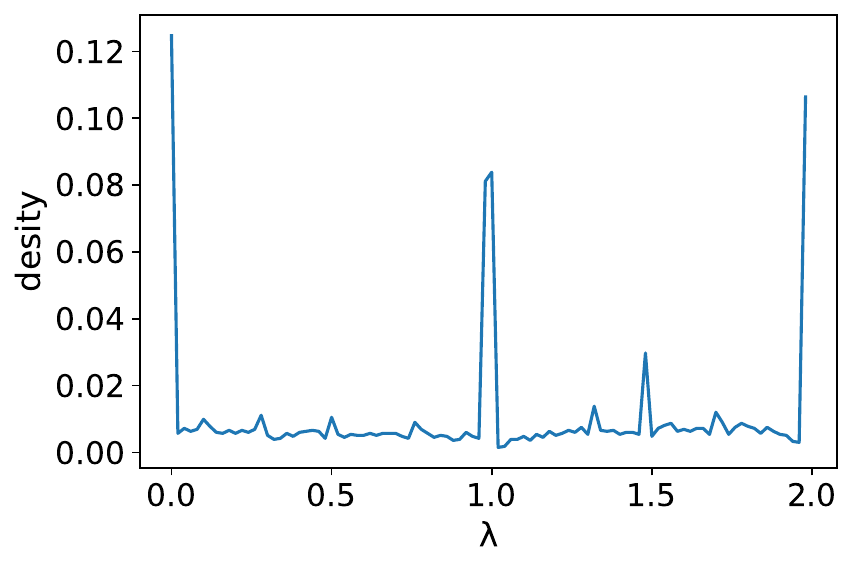}}
        \subfigure[Pubmed]{\includegraphics[width=0.48\linewidth]{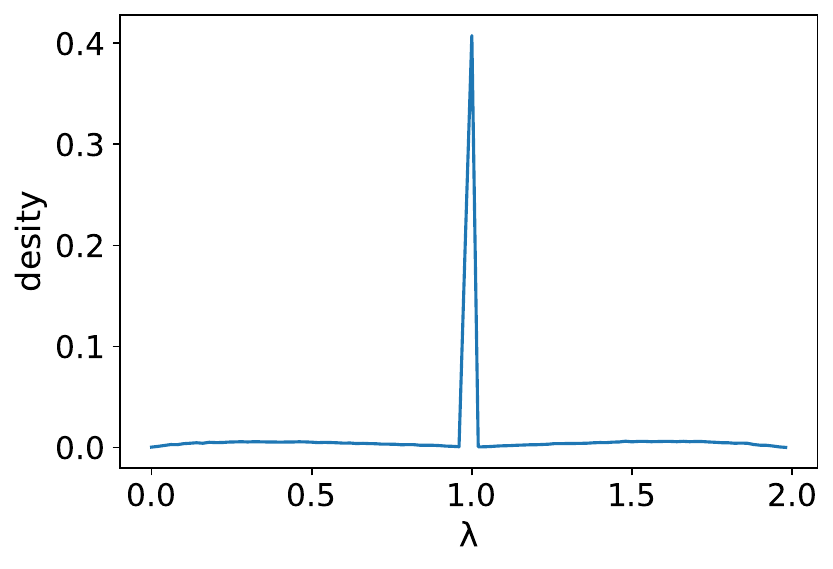}}
        \subfigure[Actor]{\includegraphics[width=0.48\linewidth]{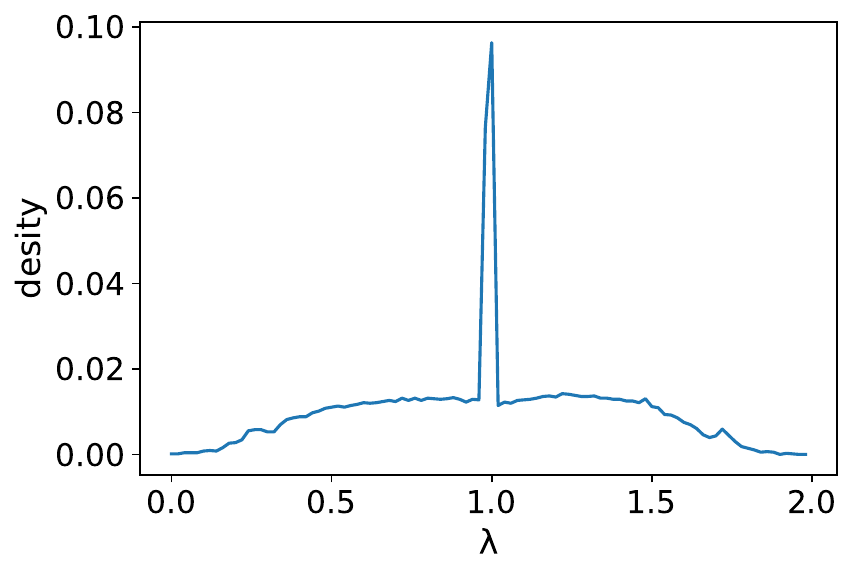}}
	\subfigure[Photo]{\includegraphics[width=0.48\linewidth]{images/photo.pdf}}
        \subfigure[Computers]{\includegraphics[width=0.48\linewidth]{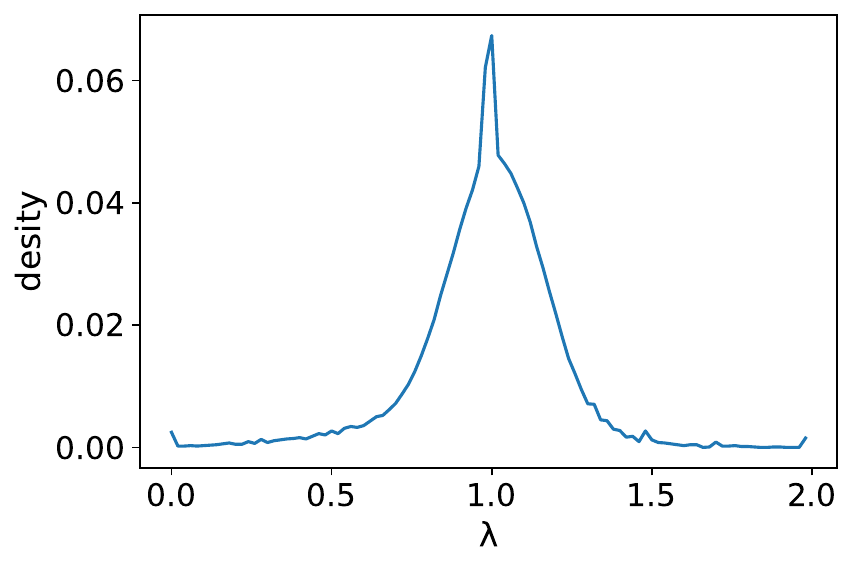}}
        \subfigure[Chameleon]{\includegraphics[width=0.48\linewidth]{images/chameleon.pdf}}
	\subfigure[Squirrel]{\includegraphics[width=0.48\linewidth]{images/squirrel.pdf}}
        \subfigure[Texas]{\includegraphics[width=0.48\linewidth]{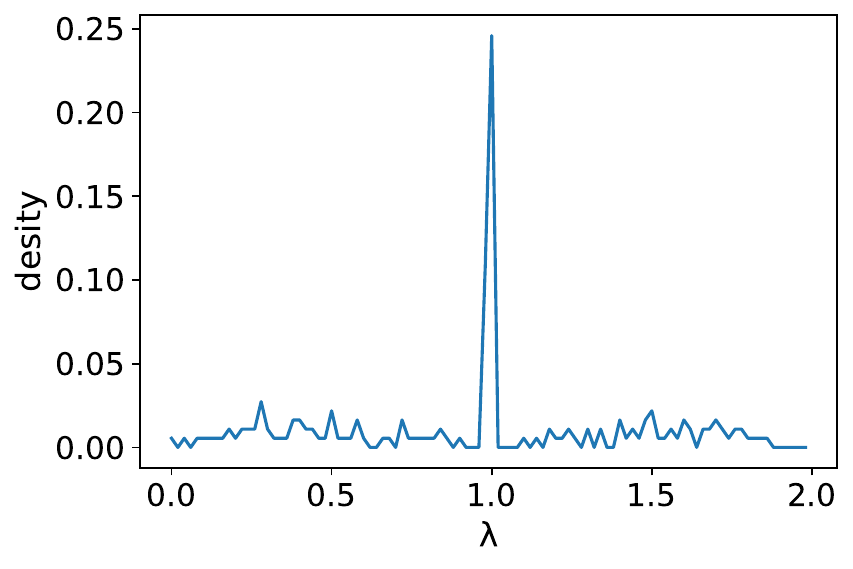}}
        \subfigure[Cornell]{\includegraphics[width=0.48\linewidth]{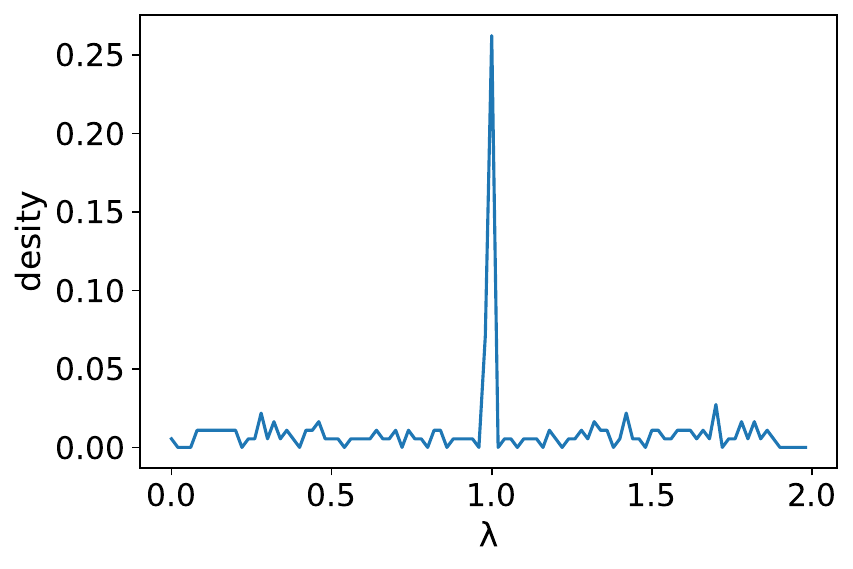}}

	\caption{Histogram of the distribution of eigenvalues of the normalized Laplacian matrix on different datasets.}
	\label{fig:histogram}
\end{figure}

\section{Eigenvalue distribution of random graphs}
Figure 7 shows the variation of eigenvalues with the average degree for a random graph with four thousand nodes. We find that the higher the average degree of the node, the more concentrated the distribution of eigenvalues is at the value 1, which is consistent with the distribution trend in Figure 1. Therefore, we speculate that the tendency that the eigenvalues of real-world graphs to be concentrated around 1 is also related to their average degree.

\begin{figure}[t] 
	\centering
 \subfigure[2]{\includegraphics[width=0.48\linewidth]{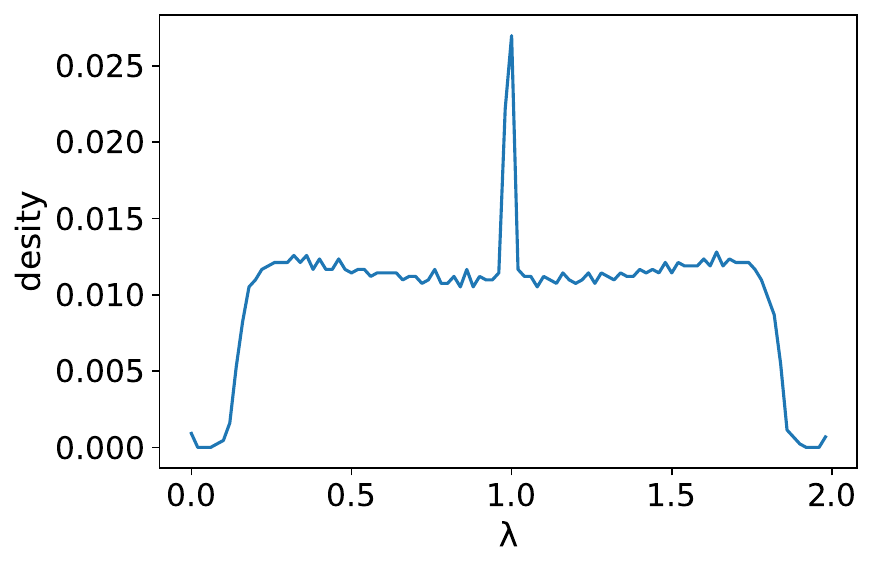}}
 \subfigure[3]{\includegraphics[width=0.48\linewidth]{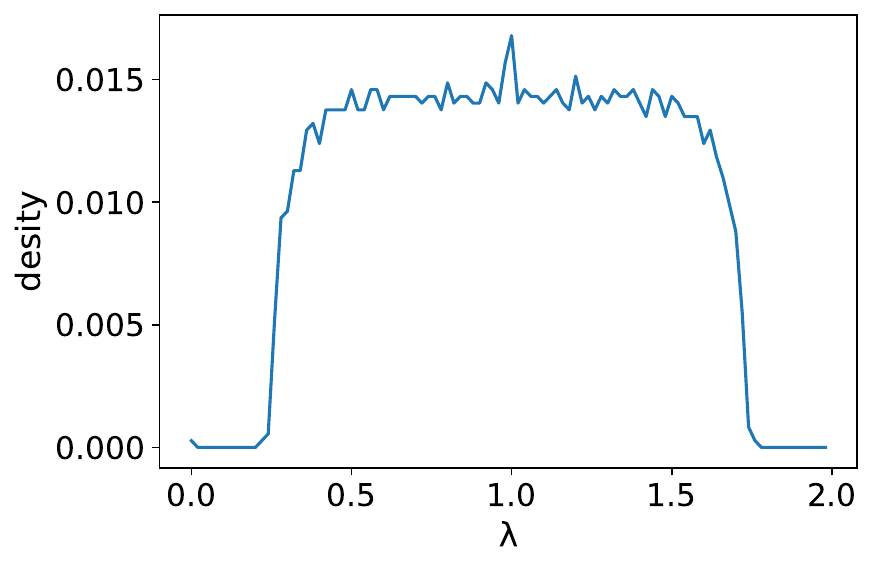}}
 \subfigure[5]{\includegraphics[width=0.48\linewidth]{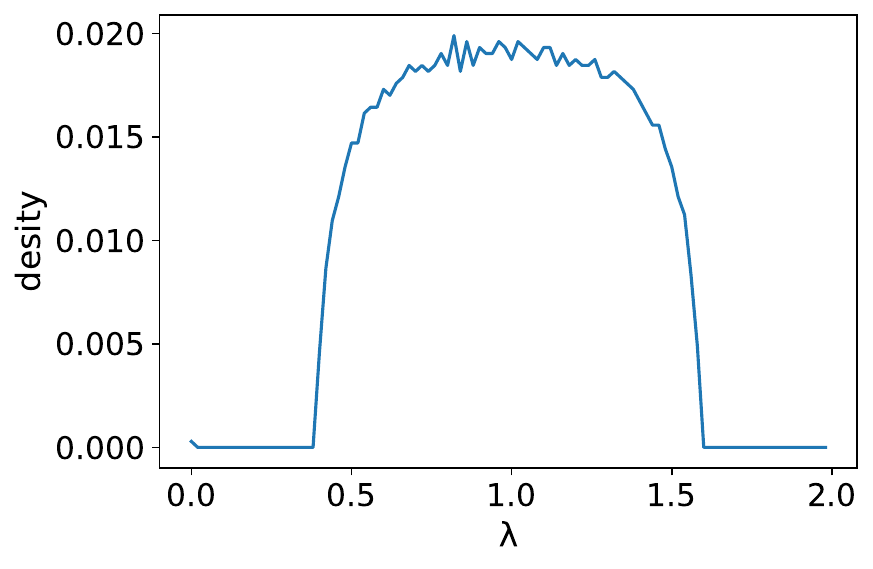}}
 \subfigure[10]{\includegraphics[width=0.48\linewidth]{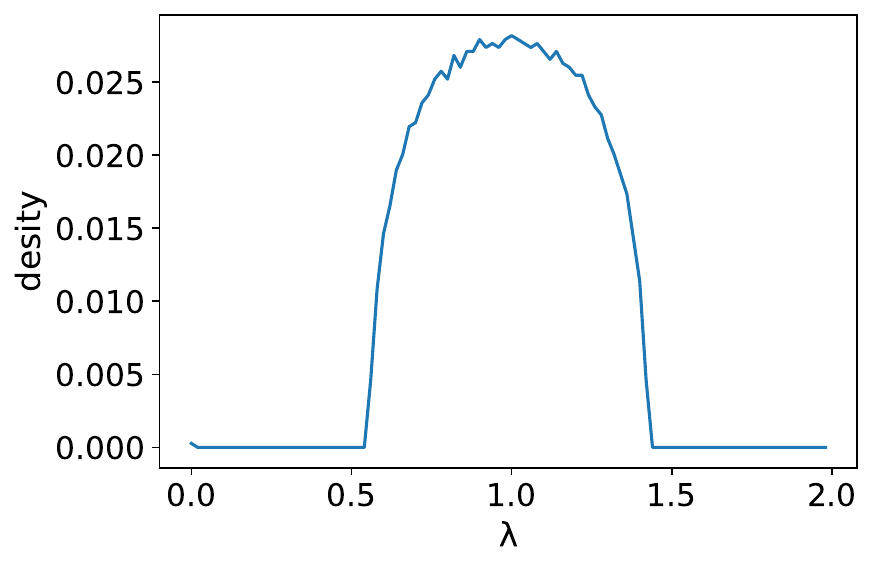}}
 \subfigure[50]{\includegraphics[width=0.48\linewidth]{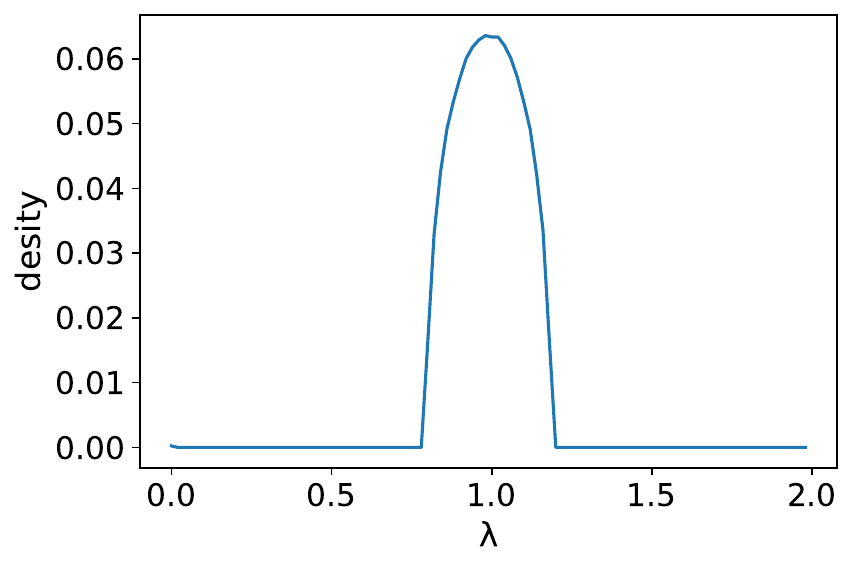}}
 \subfigure[100]{\includegraphics[width=0.48\linewidth]{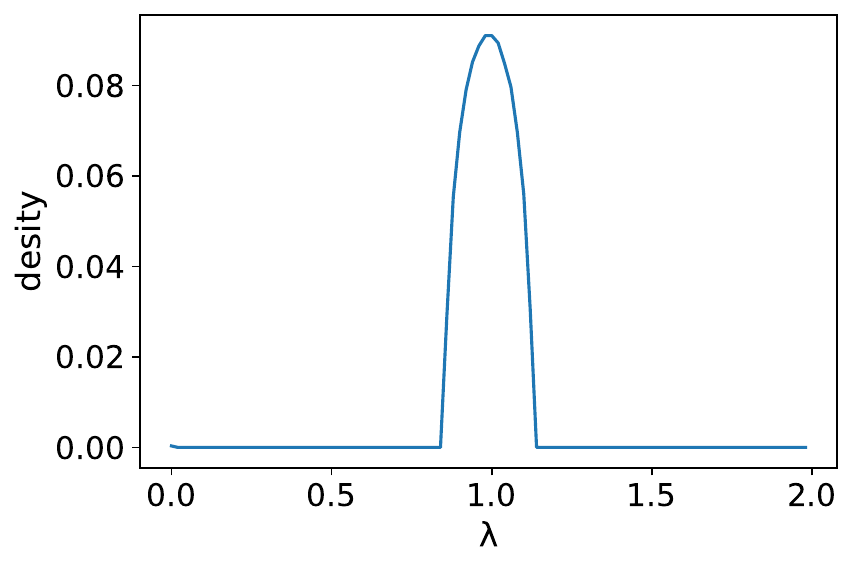}}
 \subfigure[150]{\includegraphics[width=0.48\linewidth]{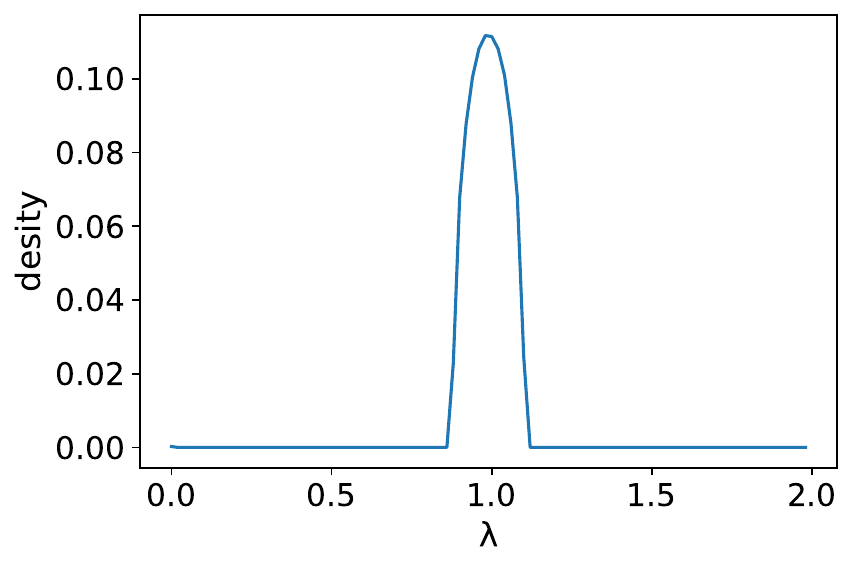}}
 \subfigure[200]{\includegraphics[width=0.48\linewidth]{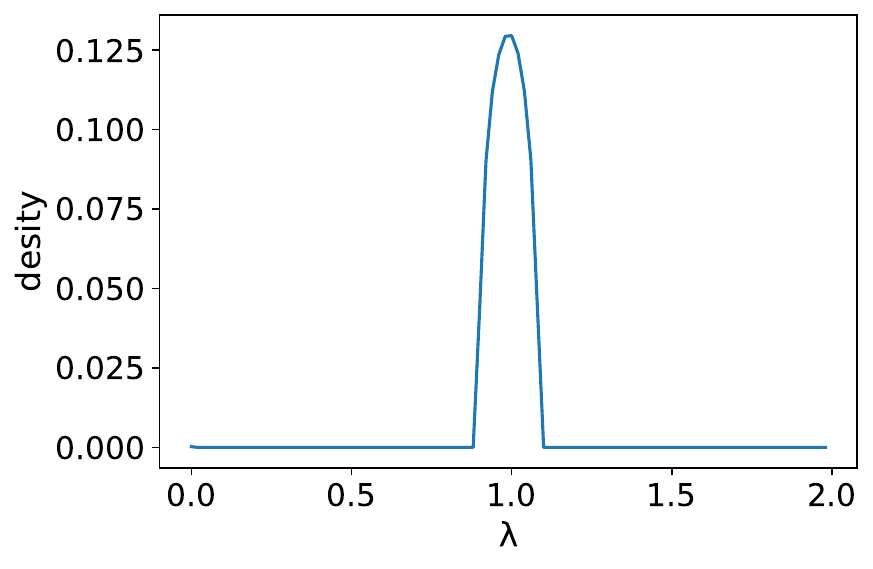}}
 \subfigure[250]{\includegraphics[width=0.48\linewidth]{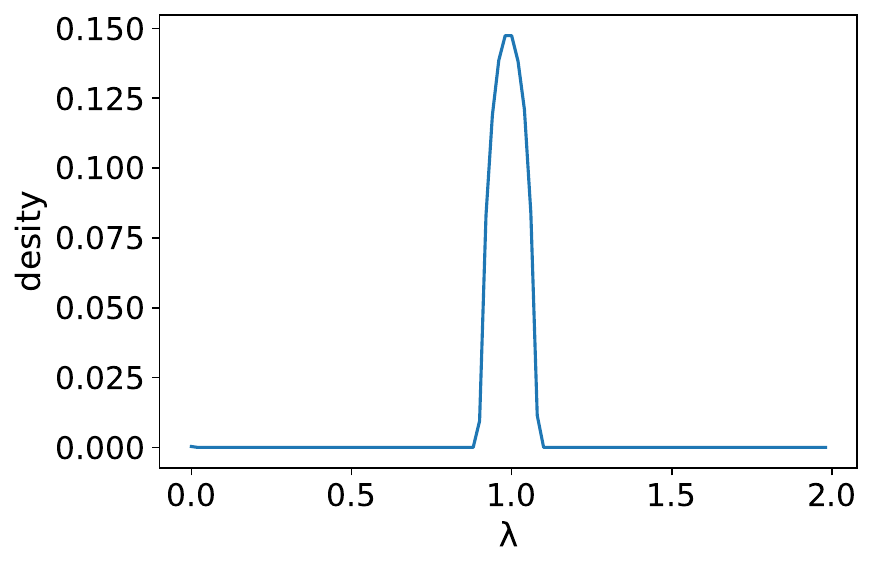}}
 \subfigure[300]{\includegraphics[width=0.48\linewidth]{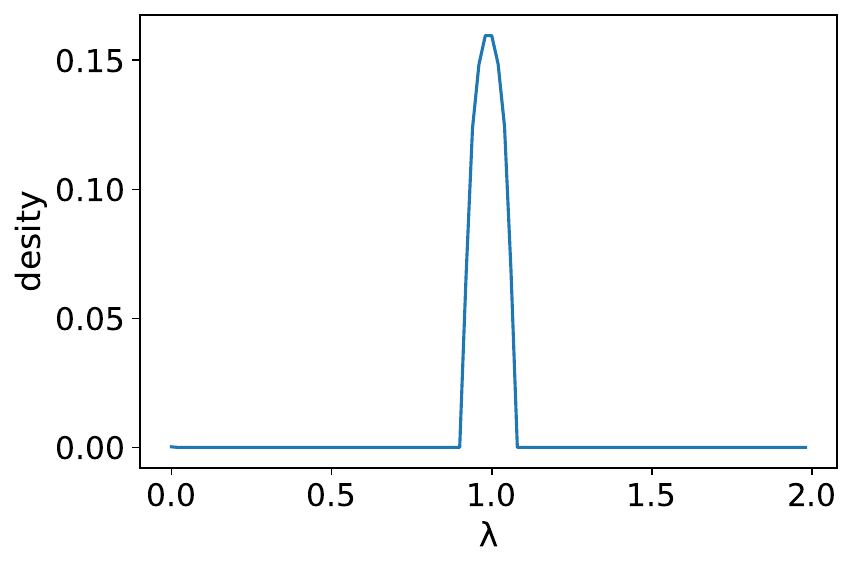}}
	\caption{The distribution of eigenvalues of random graphs varies with different average degrees.}
\end{figure}

\section{The setting of hyperparameter $\beta$}

\begin{table}[h]
\begin{tabular}{lccccc}
\hline
datasets  & Low  & High & Band & Reject & Comb \\
\hline
EC-GPR & 0.80 & 0.70 & 0.00 & 0.00   & 0.00 \\
EC-Bern & 0.60 & 0.90 & 0.70 & 0.00   & 0.00 \\
EC-Jacobi        & 0.90 & 0.90 & 0.70 & 0.70   & 0.00 \\
\hline
\end{tabular}
\caption{$\beta$-hyperparameter settings for synthetic datasets.}
\end{table}

Tables 5 and 6 present the hyperparameter $\beta$ settings of EC-GPR, EC-Bern, and EC-Jacobi on synthetic datasets and real-world datasets. 
\begin{table*}[t]
\centering
\label{tab:time}
% \vskip 0.1in
\begin{center}
\begin{small}
\resizebox{0.9\textwidth}{!}{
\setlength{\tabcolsep}{1mm}
\begin{tabular}{lcccccccccc}
\hline
datasets    & Cora & Citeseer & Pubmed & Computers & Photo & Chameleon & Actor & Squirrel & Texas & Cornell \\
\hline
EC-GPR  & 0.90 & 0.25     & 0.75   & 0.15      & 0.05  & 0.09      & 0.90  & 0.23     & 0.91  & 0.77    \\
EC-Bern & 0.63 & 0.78     & 0.05   & 0.12      & 0.05  & 0.33      & 0.86  & 0.44     & 0.86  & 0.18    \\
EC-Jacobi        & 0.50 & 0.32     & 0.50   & 0.39      & 0.47  & 0.29      & 0.26  & 0.76     & 0.40  & 0.70   \\
\hline
\end{tabular}
}
\end{small}

\caption{$\beta$-hyperparameter settings for real-world datasets.}
\end{center}
\end{table*}

\end{document}